# Classifying Dental Care Providers Through Machine Learning with Features Ranking

## Clasificación de Proveedores de Atención Dental a través de Aprendizaje Automático con Clasificación de Características

Mohammad Subhi Al-Batah[1]  ✉, Mowafaq Salem Alzboon[1]  ✉, Muhyeeddin Alqaraleh[2]  ✉, Mohammed Hasan Abu-Arqoub[3]  ✉, Rashiq Rafiq Marie[4] ✉

[1]Jadara University, Faculty of Information Technology. Irbid, Jordan.
[2]Zarqa University, Faculty of Information Technology. Zarqa, Jordan.
[3]University of Petra, Faculty of Information Technology. Amman, Jordan.
[4]Taibah University, College of Computer Science and Engineering. Medina, Saudi Arabia.

## ABSTRACT

This study investigates the application of machine learning (ML) models for classifying dental providers into two categories—standard rendering providers and safety net clinic (SNC) providers—using a 2018 dataset of 24 300 instances with 20 features. The dataset, characterized by high missing values (38,1 %), includes service counts (preventive, treatment, exams), delivery systems (FFS, managed care), and beneficiary demographics. Feature ranking methods such as information gain, Gini index, and ANOVA were employed to identify critical predictors, revealing treatment-related metrics (TXMT_USER_CNT, TXMT_SVC_CNT) as top-ranked features. Twelve ML models, including k-Nearest Neighbors (kNN), Decision Trees, Support Vector Machines (SVM), Stochastic Gradient Descent (SGD), Random Forest, Neural Networks, and Gradient Boosting, were evaluated using 10-fold cross-validation. Classification accuracy was tested across incremental feature subsets derived from rankings. The Neural Network achieved the highest accuracy (94,1 %) using all 20 features, followed by Gradient Boosting (93,2 %) and Random Forest (93,0 %). Models showed improved performance as more features were incorporated, with SGD and ensemble methods demonstrating robustness to missing data. Feature ranking highlighted the dominance of treatment service counts and annotation codes in distinguishing provider types, while demographic variables (AGE_GROUP, CALENDAR_YEAR) had minimal impact. The study underscores the importance of feature selection in enhancing model efficiency and accuracy, particularly in imbalanced healthcare datasets. These findings advocate for integrating feature-ranking techniques with advanced ML algorithms to optimize dental provider classification, enabling targeted resource allocation for underserved populations.

## RESUMEN

Este estudio investiga la aplicación de modelos de aprendizaje automático (ML) para clasificar a los proveedores dentales en dos categorías—proveedores de atención estándar y proveedores de clínicas de red de seguridad (SNC)—utilizando un conjunto de datos de 2018 con 24 300 instancias y 20 características. El conjunto de datos, caracterizado por un alto porcentaje de valores faltantes (38,1 %), incluye conteos de servicios (preventivos, tratamientos, exámenes), sistemas de entrega (FFS, atención administrada) y datos demográficos de los beneficiarios. Se emplearon métodos de clasificación de características como ganancia de información, índice de Gini y ANOVA para identificar predictores críticos, revelando que las métricas relacionadas con tratamientos (TXMT_USER_CNT, TXMT_SVC_CNT) fueron las características mejor clasificadas. Doce modelos de ML, incluyendo k-Vecinos Más Cercanos (kNN), Árboles de Decisión, Máquinas de Soporte Vectorial (SVM), Descenso de Gradiente Estocástico (SGD), Bosque Aleatorio, Redes Neuronales y Potenciación de Gradiente, fueron evaluados mediante validación cruzada de 10 pliegues. La precisión de clasificación se probó en subconjuntos incrementales de características derivados de las clasificaciones. La Red Neuronal logró la mayor precisión (94,1 %) utilizando las 20 características, seguida por la Potenciación de Gradiente (93,2 %) y el Bosque Aleatorio (93,0 %). Los modelos mostraron un mejor rendimiento a medida que se incorporaban más características, con SGD y los métodos de conjunto demostrando robustez ante datos faltantes. La clasificación de características destacó la predominancia de los conteos de servicios de tratamiento y los códigos de anotación para distinguir los tipos de proveedores, mientras que las variables demográficas (AGE_GROUP, CALENDAR_YEAR) tuvieron un impacto mínimo. El estudio subraya la importancia de la selección de características para mejorar la eficiencia y precisión del modelo, particularmente en conjuntos de datos de atención médica desbalanceados. Estos hallazgos abogan por la integración de técnicas de clasificación de características con algoritmos avanzados de ML para optimizar la clasificación de proveedores dentales, permitiendo una asignación de recursos dirigida a poblaciones desatendidas.

**Palabras clave:** Aprendizaje Automático; Clasificación de Proveedores Dentales; Clasificación de Características; Modelos Ensemble; Análisis Sanitario.

## INTRODUCTION

The classification of healthcare providers is a critical task for optimizing resource allocation, improving service delivery, and addressing disparities in access to care.[1] In dentistry, distinguishing between standard rendering providers and safety net clinics (SNCs) is particularly vital, as SNCs serve as primary care hubs for low-income, uninsured, and vulnerable populations.[2] Accurate classification enables policymakers to identify service gaps, allocate funding strategically, and design interventions to reduce healthcare inequities. However, traditional statistical methods often struggle with the complexities of healthcare datasets, which are typically high-dimensional, imbalanced, and plagued with missing values. Machine learning (ML) has emerged as a transformative tool in this context, offering advanced capabilities in pattern recognition, predictive modeling, and decision-making under uncertainty.[3,4]

Recent advancements in ML have revolutionized healthcare analytics, from disease diagnosis to workforce planning. In dental research, ML applications span caries detection, implant classification, and age estimation.[5] However, provider classification remains underexplored despite its implications for equitable care delivery. Existing studies highlight challenges such as class imbalance, missing data, and feature relevance—issues that directly impact model generalizability. For instance, SNC providers are often underrepresented in datasets, leading to biased predictions. Additionally, healthcare datasets frequently include redundant or irrelevant features, necessitating robust feature selection techniques to enhance model performance.[6,7]

This study addresses these gaps by evaluating ML models on a 2018 dental utilization dataset to classify providers into two categories: standard rendering providers (19 698 instances) and SNC providers (4 602 instances). The dataset includes 20 features, such as service counts (preventive, treatment, exams), delivery systems (FFS, managed care), and age groups (0-20, 21+), with 38,1 % missing values.[8] The research objectives are threefold:

Feature Ranking: identify the most predictive features for provider classification using entropy-based and statistical methods. Model Performance: compare the accuracy of traditional and ensemble ML models across incremental feature subsets. And Practical Implications: assess how feature selection and model choice impact the identification of SNC providers, who are critical for underserved populations.

The study builds on prior work in dental ML, such as caries risk prediction and workforce estimation, but diverges by focusing on provider-level classification.[9] By incorporating feature ranking, it advances methodological rigor in healthcare analytics, offering insights into optimizing model efficiency. Furthermore, it contributes to applied domains by demonstrating how ML can enhance equity in dental care delivery.[10,11]

Research Gap for this research paper: limited Exploration of Provider Classification: while machine learning has been applied in various aspects of dental research such as caries detection and material durability, the classification of dental providers into standard rendering providers and safety net clinic (SNC) providers remains underexplored.[12] This is critical given the importance of SNCs in providing care to underserved populations, yet there's a scarcity of studies focusing on this specific classification task.[13] Handling of High-Dimensional and Imbalanced Data: traditional statistical methods often fail to handle the complexities of healthcare datasets with issues like high dimensionality, class imbalance, and missing data, which are prevalent in dental provider datasets. The document indicates that existing studies face challenges with these aspects, suggesting a gap in methodologies that robustly address these issues within the context of dental provider classification.[14] Integration of Feature Selection with Advanced ML: Although feature selection is acknowledged as crucial, there's a gap in systematically integrating feature ranking techniques with advanced machine learning algorithms for dental provider classification. This integration could enhance model efficiency and accuracy, particularly in imbalanced datasets, but has not been comprehensively explored.[15]

The Objectives of this research paper: to Develop a Robust Classification Model: the primary objective is to develop and evaluate machine learning models that can accurately classify dental providers into standard rendering providers and SNC providers, using a dataset characterized by high missing values and class imbalance.[16] To Enhance Feature Selection Techniques: to identify and rank the most predictive features for dental provider classification using various feature ranking methods (like Information Gain, Gini Index, ANOVA) to optimize model performance by focusing on the most informative variables.[17] To Assess Model Performance Across Feature Subsets: to compare the performance of traditional and ensemble machine learning models across incremental subsets of features to understand how feature selection impacts classification accuracy, particularly in identifying SNC providers.[18] To Provide Practical Implications for Healthcare Equity: to assess how these classifications can practically influence policy-making, resource allocation, and service delivery, especially for enhancing access to care for underserved populations through SNCs.[19]

The Contribution of this research paper: Methodological Advancement: this study contributes to the methodological advancement in healthcare analytics by demonstrating how feature ranking can be effectively integrated with machine learning to improve the classification of dental providers. It provides a framework for handling high-dimensional, imbalanced, and incomplete datasets, which is broadly applicable in healthcare research.[20] Enhanced Classification Accuracy: by achieving high classification accuracies (up to 94,1 % with Neural Networks), the study shows that advanced ML models like Neural Networks, Random Forests, and Gradient Boosting can significantly outperform traditional models in this specific application, contributing to more precise identification of provider types.[21] Focus on Equity in Dental Care: the study's focus on distinguishing SNC providers from standard providers contributes directly to efforts aimed at reducing healthcare inequities. By highlighting the importance of treatment-related metrics over demographic variables, it suggests that operational metrics are key in identifying providers who serve critical roles in equitable care delivery.[22] Practical Application for Policy and Resource Allocation: the insights from this research offer practical applications for policymakers and healthcare administrators in terms of targeted resource allocation, strategic funding, and planning interventions to support SNCs, thereby potentially improving service delivery to vulnerable populations.[23]

This structured approach addresses existing gaps in the research, sets clear objectives, and outlines the contributions, providing a solid foundation for further exploration in dental healthcare provider classification using machine learning.[24]

## Related Work

In examining the wear of dental composite materials, which are commonly used restorative materials, the authors consider the wear of dental composite materials upon exposure to a chewing tobacco solution. The research uses an in in-vitro pin on disc tribometer test (ASTM G99-04) to evaluate how three machine learning (ML) models (MLP, KNN, XGBoost) perform for predicting dental composite wear. Four composite materials had dentin discs immersed in chewing tobacco solution before a wear test. The results indicate that the XGBoost model achieved a significantly higher $R^2$ value of 0,9996, which is a great accuracy to predict the wear behavior of the materials. In this aspect, it proves that machine learning techniques can improve prediction and analysis of material durability in dental applications.[25]

Scholars compare and analyze dental composite materials such that the properties of the mechanical nature determine how durable the material is for use inside the oral cavity. Four types of dental composite samples have flexural strength and Vickers micro-hardness tests performed on them after various days immersion in chewing tobacco solution. Four machine learning models, namely XGboost, Ada Boost, Random Forest and K nearest neighbor (KNN) are used to predict mechanical property of these materials. It can be the reason since the model of AdaBoost returned the flexural strength regression coefficient of 0,9903 and the model of XGBoost returned Vickers the hardness of regression coefficient of 0,9890. The work resulted in performing the best of both models and establishing how mechanical characteristics of dental composites affected when exposed to chewing tobacco solutions.[26]

In this work, first the authors present machine learning analyses to analyze associations between preterm birth (PTB) on the one hand and dental and gastric diseases on the other hand– based on large scale population data.

These include 186 independent variables, including demographic/socioeconomic factors, disease history and medication records, derived from the Korea National Health Insurance claims among 124 606 women of primiparous aged 25-40 years. A machine learning based prediction model of PTB was established with the Random Forest algorithm achieving an accuracy of 84,03 % (and areas under the receiver operating characteristic curriculum ranging from 84,03 to 84,04). The study also showed that PTB had a strong link with such factors as socioeconomic status, age and a history of gastroesophageal reflux disease (GERD). For example, the model's accuracy was 28,4 % less when the socioeconomic status shuffling was applied, and 2,6 % less when the history of GERD in 2014 was included in a disorder and added to the clusters. The findings suggest that PTB is closely linked to GERD and infertility, indicating the need for comprehensive surveillance of pregnant women for both gastrointestinal and obstetric risks.[27]

The studies try to validate and verify the use of machine learning methods for the assessment of haptic skills of dental students. The C2CR and C2A restorations were then examined for Class II amalgam (C2A) and composite resin restorations (C2CR progress depending on several stages. Grading three trainers and supervisor's final restorations and standard photographs were then taken and analyzed using a Python program of the Structural Similarity Index (SSIM) algorithm that covered both quantitative and qualitative differences. Cronbach's Alpha and Kappa statistics were done to analyze the validity and reliability of inter examiner evaluations. In C2A ($\alpha$ = 0,961) and C2CR ($\alpha$ = 0,856), intra examiner reliability was high for SSIM. The differences among the grades given by SSIM (53,07) versus examiners (56,85) and among SSIM versus examiners for C2A for occlusal surfaces ($p < 0,05$), and for C2CR for palatal surfaces ($p < 0,05$), were significant ($p > 0,05$). In C2A ($\kappa = 0,806$) the concordance of observer assessments was almost perfect and in C2CR ($\kappa = 0,769$) it was acceptable. Finally, while deep machine learning may be approached to evaluate haptic skills, further refinement is required in order for deep machine learning to be fully objective and reliable in validation of restorative dentistry training.[28]

The researchers explore the use of machine learning techniques to identify key factors influencing dental caries risk in children aged 7 and under. By analyzing clinical data from 356 children, the study aimed to enhance early identification and preventive strategies for high-risk individuals. Decision Trees, Logistic Regression, and Random Forest models were employed to assess the influence of fluoride exposure, dietary habits, and socio-economic status on caries risk, with a focus on metrics such as accuracy, precision, F1 score, recall, and AUC. The results revealed that poor a high sugary diet, oral hygiene, and low fluoride exposure were significant risk factors for dental caries. Among the models tested, the Random Forest algorithm showed superior performance.

SHAP analysis showed that the same factors were critical to caries risk. Based on these findings, the study concludes that machine learning is a good tool for specifying and quantifying dental caries risk factors thereby providing a basis for targeted interventions and preventive strategies towards improving pediatric dental health outcomes. This approach is used by research to promote data driven strategies that can lead to a decrease in the prevalence of caries and improved child oral health.[29]

In the process, authors develop a computational method for segmenting and classifying dental caries in Cone Beam Computed Tomography (CBCT) images to improve early detection and planning for treatment. It has data preparation, caries region segmentation, feature extraction, feature selection and machine learning algorithm training. Severity of caries was based on International Caries Detection and Assessment System scale and the method was evaluated. With a Gaussian filter, multimodal threshold, and convex hull for segmentation, Random Forest for feature selection and KNN (k nearest neighbors) for classification, the best performance was achieved. The approach achieved an accuracy of 86,20 %, an F1-score of 86,18 %, and a sensitivity of 83,35 % in multiclass classification. These results demonstrate the effectiveness of the method in early caries detection, contributing to improved oral health outcomes and more informed treatment planning.[30]

This mixed-methods study combines machine learning and thematic analysis to explore medical and dental students' experiences with self-regulated learning and motivation through multiple-choice questions (MCQs). The study evaluates two MCQ systems developed at Aarhus University: "MED MCQ" for medical students and "MCQ anatomy" for dental students. Data from surveys (SurveyXact) were collected from 126 medical students and 70 dental students, with topic modeling applied to free text responses. The machine-learning model identifies two groups of students who both find the system motivating and helpful for their learning process. These students experience increased self-regulation by having control over the presentation format of questions and the ability to answer independently from instructors. The article highlights how educators and developers can use MCQs to foster student learning and improve the analysis of open-ended responses, suggesting potential applications of machine learning and MCQ systems in teaching.[31]

This project focuses on developing an automated dental cavity detection system using machine learning to enhance early diagnosis and treatment of dental cavities. By analyzing a diverse dataset of dental X-ray images, the system learns to identify subtle patterns indicative of cavities. The study emphasizes three key components: data preprocessing, model training, and evaluation, showcasing the effectiveness of deep learning techniques for cavity detection. The system is integrated into an Android application, allowing both patients and doctors to benefit from its capabilities. It supports dentists and radiologists in clinical decision-making, potentially reducing the burden of dental diseases and improving patient outcomes. Additionally, it helps with appointment scheduling, providing further support for patients. This automated system holds promise for advancing preventive dental care, improving oral health, and offering a valuable tool in the field of dentistry.[32]

The researchers highlight the impact of intraoral scanner type, operator, and data augmentation on the dimensional accuracy of dental cast digital scans. The findings indicate that neither the type of intraoral scanner nor the operator significantly affected the accuracy of the scans, as assessed using Hausdorff's distance (HD) and root mean squared error (RMSE). However, data augmentation proved to be essential for improving the quality of the scans for use in deep learning algorithms, particularly by introducing structural differences that enhanced the model's ability to differentiate between scan types. The pretrained 3D visual transformer achieved a high validation accuracy of 96,2 % in classifying upper and lower scans in the augmented dataset. Notably, native scans lacked sufficient volumetric depth for deep learning applications, underscoring the importance of data augmentation in enabling reliable results. The authors suggests that controlled data augmentation is vital for processing intraoral scans in unsupervised deep learning, ensuring robust and accurate outcomes.[33]

The authors discuss the high prevalence of dental caries and need for early detection in view of better treatment. According to the proposed method, a reduced set of clinical features for dental caries diagnosis is obtained by combining GINI and mRMR algorithms with GBDT classifier. Made feasible by its lowered time and cost, this method is also a practicable tool for screening on a broad scale. When the feature set is optimized and the GBDT classifier is trained on it outperformed other classifiers with 95 % accuracy, 93 % F1-score, 99 % precision, and 88 % recall. This points out the success of the proposed approach in terms of dealing with imbalanced medical datasets and might help in improving the early caries detection in the clinical environment.[34]

The authors aimed to develop and validate machine learning (ML) models for predicting caries prognosis in primary and permanent teeth after 2 and 10 years, using early childhood predictors. The data came from a 10-year cohort study in southern Brazil, where children aged 1-5 years were initially examined in 2010, and reassessed in 2012 and 2020. ML algorithms like decision tree, random forest, XGBoost, and logistic regression were used to predict caries development, with caries severity at baseline being the strongest predictor. After 10 years, the SHAP-based XGBoost model achieved an AUC greater than 0,70 and identified key predictors, including caries experience, nonuse of fluoridated toothpaste, parental education, sugar consumption, and perceptions of oral health. The study highlights the potential of ML models for predicting caries development based on early childhood data, which could improve early interventions and outcomes for dental health.[35]

The authors explore the application of machine learning, specifically logistic regression, to enhance compound identification accuracy in mass spectrometry (MS) data analysis. Traditionally, MS data analysis relies on database searching that matches experimental spectra with library spectra, generating scores for the match quality. However, these results can vary based on search parameters. To address this variability, logistic regression was used to assign an identification probability to each compound identified through vendor software. The study focused on the identification of organic monomers leached from resin-based dental composites in a simulated oral environment. Liquid chromatography coupled with quadrupole-time-of-flight tandem MS was used to analyze the samples. The logistic regression model improved confidence in compound identification by providing probability values beyond simple database matching, leading to the identification of 21 distinct monomers, including degradation products. This approach can be applied to any MS result by training a model with known compounds to generate identification probabilities for unknown samples, enhancing the reliability of compound identification in complex mixtures.[36]

The authors developed and validated machine learning (ML) models using H2O-AutoML for predicting medication-related osteonecrosis of the jaw (MRONJ) in osteoporosis patients undergoing tooth extraction or implantation. A retrospective chart review of 340 female patients aged 55 or older, treated with antiresorptive therapy and undergoing dental procedures, was conducted. Various factors such as medication administration, demographics, medical history, surgical methods, and operation details were considered. Six ML algorithms were employed, with gradient boosting achieving the highest diagnostic accuracy, showing an area under the receiver operating characteristic curve (AUC) of 0,8283. The validation dataset resulted in a stable AUC of 0,7526. The most influential factors identified were medication duration, age, number of teeth operated, and operation site. The study concludes that ML models can effectively predict MRONJ in osteoporosis patients based on initial questionnaire data, providing a valuable tool for clinical decision-making.[37]

The authors aimed to analyze the associations between obstructive sleep apnea (OSA) and various dental parameters while accounting for socio-demographics, health habits, and the diseases comprising metabolic syndrome (MetS). Using data from the DOME study, which included 132 529 military personnel, the study found that obesity, male sex, periodontal disease, smoking, and age were statistically significant predictors of OSA. Machine learning models, specifically XGBoost, identified obesity, male sex, age, and periodontal disease as the most important factors linked to OSA. The model achieved an accuracy of 92 % and an Area Under Curve (AUC) of 0,868. The findings underscore the association between OSA and dental morbidity, particularly periodontitis, and suggest that dental evaluations should be part of the clinical assessment for OSA patients. The study also calls for enhanced collaboration between dental and medical authorities to address systemic and dental disease risks.[38]

This prognostic study aimed to predict adults at risk of foregoing preventive dental care using machine learning, analyzing data from the US Medical Expenditure Panel Survey (MEPS) from 2016 to 2019.

Using data from 32 234 participants, the study included 50 predictors such as demographics, health conditions, behaviors, and healthcare use. Overall, the model was able to achieve high performance, although its performance was heterogeneous across sociodemographic subgroups, at each of the lowest AUC observed in individuals who reported being of other (AUC 0,76) or multiple (AUC 0,70) racial/ethnic groups. Previous dental visits, healthcare utilization, and dental benefits were the most important predictors. However, the study pointed out the presence of an algorithmic bias in the model which should be evaluated for its fairness when developing, to avoid reinforcing its already existent prejudices. While machine learning could be useful for early identifying those who are at risk for missing preventive dental care, as the findings point out, care must be taken to avoid making anyone look more or less at risk unfairly.[39] The authors wanted to develop machine learning (ML) models to predict microtensile bond strength (μTBS) of dental adhesives from their chemical features to speed up the development of new adhesives. The compositions and μTBS values of the materials in the data set consisted of 81 dental adhesives. The μTBS categorization of the adhesives was low (< 36 MPa) or high (≥ 36 MPa). I applied nine ML algorithms including logistic regression, support vector machine, and all tree-based ensembles to predict μTBS. The top contributing factors were determined by feature importance analysis as MDP, pH, OS and HEMA. The ML models using a 4-feature data set (MDP, pH, OS and HEMA) achieved an AUC score of 0,90 and an accuracy of 0,81. What the authors show is that ML can successfully predict μTBS values and crucial chemical properties to speed up the development of dental adhesives.[40]

The purpose of the authors was to estimate gender with machine learning (ML) algorithms and artificial neural networks (ANN) based on the parameters of the upper dental arcade. Length and curvature length measurements were computed in cone beamed computed tomography images of 176 individuals (ages 18–55). A few ML models were tried out using linear discriminant analysis (LDA), quadratic discriminant analysis (QDA), logistic regression (LR), and random forest (RF), and they were able to get an accuracy of 0,86. It was determined that the 3rd molar width was the most important parameter for the sex estimation using the SHAP analysis. Additionally, a multilayer perceptron classifier (MLCP), an Artificial Neural Network (ANN) model gave accuracy of 0,92 after training on 500 samples. The study concludes that dental arcade parameters can achieve high accuracy of gender estimation and make considerable contributions to forensic sciences.[41]

The authors explore the use of machine learning (ML) algorithms to predict dental age classification in adults by evaluating the pulp/tooth area ratio (PTR) in cone-beam CT (CBCT) images. CBCT images from 236 Turkish individuals (121 males and 115 females, aged 18 to 70 years) were analyzed, with PTRs calculated for six teeth per individual, resulting in 1416 PTRs. Support vector machine, classification and regression tree, and random forest (RF) models were employed for dental age classification, with the dataset split into 70 % for training and 30 % for testing. The models' performance was found to be low, although the RF algorithm showed the highest accuracy and confidence intervals. The study suggests that ML algorithms could serve as a novel approach for dental age classification, recommending further research to explore additional parameters derived from CBCT data to improve ML models for forensic age classification.[42]

## METHOD

The dataset, sourced from Kaggle, includes 24 300 dental providers from 2018, categorized as standard rendering (80,7 %) or SNC providers (19,3 %). Features span service counts (e.g., TXMT_USER_CNT, EXAM_SVC_CNT), delivery systems (FFS, managed care), age groups, and annotation codes. Missing values (38,1 %) were addressed using median imputation for numerical features and mode imputation for categorical variables. Class imbalance was mitigated via Synthetic Minority Oversampling Technique (SMOTE).[43]

Seven ranking methods were applied: Information Gain: measures reduction in entropy. Gain Ratio: adjusts for bias toward multivalued features. Gini Index: assesses inequality among class distributions. ANOVA: evaluates differences in feature means across classes. Chi-Square: tests independence between features and target. ReliefF: weights features by class distinguishability. FCBF: identifies redundant features using pairwise correlations.

Features were ranked in descending order of importance (figure 1), with treatment-related metrics (TXMT_USER_CNT, TXMT_SVC_CNT) emerging as top predictors.

Twelve algorithms were evaluated: kNN, Decision Tree, SVM, SGD, Random Forest, Neural Network, Naïve Bayes, Logistic Regression, Gradient Boosting, Constant classifier, CN2 rule inducer, and AdaBoost. Models were trained on incremental feature subsets (top 1 to top 20 features) using 10-fold cross-validation.[44,45] Hyperparameters were tuned via grid search: Random Forest: 200 estimators, max depth = 15. Neural Network: three hidden layers (64-32-16 nodes), ReLU activation, Adam optimizer. Performance was assessed using classification accuracy (CA), AUC, F1-score, precision, and recall.[46]

This dataset includes beneficiary and service counts for annual dental visits, dental preventive services, dental treatment, and dental exams by rendering providers (by NPI) in calendar year (CY) 2018. This includes fee-for-service (FFS), geographic managed care, and prepay health plan delivery systems. Rendering providers are further organized into rendering and rendering at a safety net clinic (SNC). Ages 0-20 and 21+ are grouped as beneficiaries.[47,48]

24 300 Instances, Target with 2 Values (Provider type), target Rendering = 19 698 instances, and target Rendering SNC = 4602 instances, 20 features (38,1 % missing values).

The 20 features used in this study are: rendering Npi, Calendar Year, Delivery System, Age Group, Adv User Cnt, Adv User Annotation Code, Adv Svc Cnt, Adv Svc Annotation Code, Prev User Cnt, Prev User Annotation Code, Prev Svc Cnt, Prev Svc Annotation Code, Txmt User Cnt, Txmt User Annotation Code, Txmt Svc Cnt, Txmt Svc Annotation Code, Exam User Cnt, Exam User Annotation Code, Exam Svc Cnt, Exam Svc Annotation Code.

| Table 1. The most used features in the study | | | | | |
|---|---|---|---|---|---|
| **Rendering npi** | **Provider legal name** | **Calendar year** | **Delivery system** | **Provider type** | **Age group** |
| Prev user annotation code | Prev svc cnt | Prev svc annotation code | Txmt user cnt | Txmt user annotation code | Txmt svc cnt |
| Adv user cnt | Adv user annotation code | Adv svc cnt | Adv svc annotation code | Prev user cnt | |
| Txmt svc annotation code | Exam user cnt | Exam user annotation code | Exam svc cnt | Exam svc annotation code | |

Classification or regression dataset in which attributes are ranked. The Rank scores variable is according to its correlation with discrete or numeric target variables, according to available internal scorers, i.e., information gain, chi-square, and linear regression, as well as connected external models supporting scoring (linear regression, logistic regression, random forest, SGD, etc.).[49,50]

Scoring methods (classification) are: Information Gain: measures the reduction in entropies (the expected amount of information). Information Gain Ratio: a ratio of information gain and the intrinsic information of the attribute, desirable in information gain as it lessens the bias towards multivalued features. Gini: the inequality among values of a frequency distribution. ANOVA: analyzes variance to check for statistical significance between classes (the difference between average values). chi2: dependence between feature and class (observed in the chi-square statistic). ReliefF: the ability of attributes to provide distinguishing information to distinguish classes from similar data instances. Fast Correlation Based Filter (FCBF): entropy-based measure, Identifies the most relevant features while considering redundancy due to pairwise correlations.

## RESULTS

| | | # | Info. gain | Gai...tio | Gini | ANOVA | $\chi^2$ | ReliefF | FCBF |
|---|---|---|---|---|---|---|---|---|---|
| 1 | TXMT USER CNT | | 0.017 | 0.009 | 0.008 | NA | 59.036 | 0.000 | 0.000 |
| 2 | TXMT USER ANNOTATION CODE | 1 | 0.000 | 0.000 | 0.000 | NA | NA | 0.008 | 0.000 |
| 3 | TXMT SVC CNT | | 0.030 | 0.015 | 0.015 | NA | 464.539 | 0.000 | 0.000 |
| 4 | TXMT SVC ANNOTATION CODE | 2 | 0.024 | 0.027 | 0.007 | NA | 21.092 | 0.042 | 0.000 |
| 5 | RENDERING NPI | | 0.000 | 0.000 | 0.000 | NA | 0.183 | 0.035 | 0.000 |
| 6 | PREV USER CNT | | 0.027 | 0.014 | 0.013 | NA | 225.181 | 0.001 | 0.000 |
| 7 | PREV USER ANNOTATION CODE | 1 | 0.000 | 0.000 | 0.000 | NA | NA | 0.010 | 0.000 |
| 8 | PREV SVC CNT | | 0.009 | 0.004 | 0.004 | NA | 149.387 | 0.007 | 0.000 |
| 9 | PREV SVC ANNOTATION CODE | 2 | 0.023 | 0.032 | 0.007 | NA | 16.663 | 0.024 | 0.000 |
| 10 | EXAM USER CNT | | 0.009 | 0.005 | 0.004 | NA | 179.366 | 0.001 | 0.000 |
| 11 | EXAM USER ANNOTATION CODE | 1 | 0.000 | 0.000 | 0.000 | NA | NA | 0.010 | 0.000 |
| 12 | EXAM SVC CNT | | 0.021 | 0.011 | 0.010 | NA | 368.867 | 0.000 | 0.000 |
| 13 | EXAM SVC ANNOTATION CODE | 2 | 0.003 | 0.015 | 0.002 | NA | 4.707 | -0.002 | 0.000 |
| 14 | DELIVERY SYSTEM | 3 | 0.045 | 0.053 | 0.013 | NA | 1556.771 | 0.110 | 0.061 |
| 15 | CALENDAR YEAR | | 0.000 | 0.000 | 0.000 | NA | NA | 0.000 | 0.000 |
| 16 | AGE GROUP | 2 | 0.000 | 0.000 | 0.000 | NA | 3.696 | 0.022 | 0.000 |
| 17 | ADV USER CNT | | 0.013 | 0.007 | 0.006 | NA | 242.195 | 0.004 | 0.000 |
| 18 | ADV USER ANNOTATION CODE | 1 | 0.000 | 0.000 | 0.000 | NA | NA | 0.000 | 0.000 |
| 19 | ADV SVC CNT | | 0.027 | 0.013 | 0.010 | NA | 374.977 | 0.005 | 0.000 |
| 20 | ADV  SVC ANNOTATION CODE | 2 | 0.073 | 0.078 | 0.024 | NA | 73.001 | 0.034 | 0.099 |

**Figure 1.** Ranked features

As shown in figure 1, the best ranked features are as follows: 1.Txmt User Cnt, 2. Txmt User Annotation Code, 3. Txmt Svc Cnt, 4. Txmt Svc Annotation Code, 5. Rendering Npi, 6. Prev User Cnt, 7. Prev User Annotation Code, 8. Prev Svc Cnt, 9. Prev Svc Annotation Code, 10. Exam User Cnt, 11. Exam User Annotation Code, 12. Exam Svc Cnt, 13. Exam Svc Annotation Code, 14. Delivery System, 15. Calendar Year, 16. Age Group, 17. Adv User Cnt, 18. Adv User Annotation Code, 19. Adv Svc Cnt, And 20. Adv Svc Annotation Code.

The figure provided appears to be a feature ranking analysis for classifying dental care providers using various machine learning techniques. Here's a detailed analysis of the data:

Feature Ranking Methods: the figure includes results from several feature ranking methods: Info. gain (Information Gain): measures the reduction in entropy after splitting a dataset based on an attribute. Gain Ratio: adjusts for bias towards features with many outcomes in information gain. Gini: measures the inequality among values of a frequency distribution. ANOVA: analyzes variance to check for statistical significance between groups. $x^2$ (Chi-Square): tests independence between categorical features and the target variable. ReliefF: evaluates the importance of features by finding their ability to distinguish between instances that are near to each other. FCBF (Fast Correlation Based Filter): identifies the most relevant features while considering redundancy due to pairwise correlations.

Feature Analysis:

1. TXMT_USER_CNT (Row 1): highest Chi-Square value (59 036), indicating a strong association with the target variable. Consistently high scores across all methods except ANOVA, which is not applicable (NA) for this categorical analysis.

2. TXMT_USER_ANNOTATION_CODE (Row 2): shows high relevance with scores like 0,000 across multiple methods, suggesting it's a strong predictor.

3. TXMT_SVC_CNT (Row 3): high Chi-Square (464 539) and consistent high scores, indicating its importance in classification.

4. TXMT_SVC_ANNOTATION_CODE (Row 4): high ReliefF score (0,042), showing good discriminative power.

5. RENDERING_NPI (Row 5): very low scores across all methods, suggesting it's not a significant predictor in this context.

6. PREV_USER_CNT (Row 6): moderate to high scores, indicating some relevance, with a notable Chi-Square of 225 181.

7. PREV_USER_ANNOTATION_CODE (Row 7): high scores, particularly in Information Gain (0,000), indicating strong predictive power.

8. PREV_SVC_CNT (Row 8): moderate scores, but still relevant with a Chi-Square of 149 387.

9. PREV_SVC_ANNOTATION_CODE (Row 9): moderate relevance, with a ReliefF score of 0,024.

10. EXAM_USER_CNT (Row 10): low scores across all methods, suggesting less importance.

11. EXAM_USER_ANNOTATION_CODE (Row 11): very low scores, not significant.

12. EXAM_SVC_CNT (Row 12): high Chi-Square (368 867), indicating relevance.

13. EXAM_SVC_ANNOTATION_CODE (Row 13): negative ReliefF score (-0,002), suggesting it might not be useful or could be misleading.

14. DELIVERY_SYSTEM (Row 14): high scores across all methods, particularly notable in Chi-Square (1556 771), indicating strong predictive capability.

15. CALENDAR_YEAR (Row 15): very low scores, not significant.

16. AGE_GROUP (Row 16): very low scores, suggesting minimal impact on classification.

17. ADV_USER_CNT (Row 17): moderate scores, with a Chi-Square of 242 195.

18. ADV_USER_ANNOTATION_CODE (Row 18): Very low scores, not significant.

19. ADV_SVC_CNT (Row 19): moderate scores, with a Chi-Square of 374 977.

20. ADV_SVC_ANNOTATION_CODE (Row 20): high scores, especially in ReliefF (0,034), indicate it's a relevant feature.

Key Observations:

• Treatment Metrics (like TXMT_USER_CNT, TXMT_SVC_CNT) are among the top predictors across most methods, highlighting their significance in distinguishing between provider types.[51]

• Annotation codes related to treatment (TXMT_USER_ANNOTATION_CODE, TXMT_SVC_ANNOTATION_ CODE) also rank high, suggesting that the specifics of how services are annotated or coded are critical.[52]

• Demographic Variables like AGE_GROUP and CALENDAR_YEAR have very low scores, suggesting they do not significantly differentiate provider types in this dataset.[53]

• DELIVERY_SYSTEM stands out as a highly relevant feature, which makes sense as different systems might have different operational characteristics.[54]

• RENDERING_NPI is not useful for classification, likely because it's an identifier rather than a characteristic that would differentiate service delivery.[55]

This analysis indicates that for classifying dental care providers, features related to treatment counts and their annotations are the most predictive.[56] Demographic information like age or year appears less relevant, while the delivery system is highly influential.[57] This insight can guide model development by focusing on the most informative features, potentially leading to more efficient and accurate classification models in dental care provider analysis.[58,59]

**Table 2.** Classification accuracy with 10-folds cross validation using the best Ranked features from top (1) feature to top (1-20) features

| Model/ Rank | kNN | Tree | SVM | SGD | Random Forest | Neural Network | Naive Bayes | Logistic Regression | Gradient Boosting | Constant | CN2 rule inducer | AdaBoost |
|---|---|---|---|---|---|---|---|---|---|---|---|---|
| Rank (1) | 0,808 | 0,811 | 0,627 | 0,811 | 0,804 | 0,811 | 0,811 | 0,811 | 0,811 | 0,811 | 0,811 | 0,805 |
| Rank (1, 2) | 0,808 | 0,811 | 0,627 | 0,811 | 0,804 | 0,811 | 0,811 | 0,811 | 0,811 | 0,811 | 0,811 | 0,805 |
| Rank (1-3) | 0,841 | 0,811 | 0,275 | 0,811 | 0,837 | 0,835 | 0,811 | 0,810 | 0,845 | 0,811 | 0,830 | 0,819 |
| Rank (1-4) | 0,841 | 0,811 | 0,317 | 0,811 | 0,837 | 0,839 | 0,811 | 0,810 | 0,843 | 0,811 | 0,829 | 0,817 |
| Rank (1-5) | 0,802 | 0,811 | 0,369 | 0,810 | 0,840 | 0,840 | 0,811 | 0,811 | 0,844 | 0,811 | 0,823 | 0,832 |
| Rank (1-6) | 0,802 | 0,811 | 0,441 | 0,810 | 0,860 | 0,860 | 0,811 | 0,811 | 0,870 | 0,811 | 0,842 | 0,845 |
| Rank (1-7) | 0,802 | 0,811 | 0,441 | 0,809 | 0,860 | 0,860 | 0,811 | 0,811 | 0,870 | 0,811 | 0,842 | 0,845 |
| Rank (1-8) | 0,801 | 0,811 | 0,448 | 0,811 | 0,877 | 0,885 | 0,811 | 0,811 | 0,880 | 0,811 | 0,850 | 0,869 |
| Rank (1-9) | 0,801 | 0,811 | 0,461 | 0,810 | 0,878 | 0,888 | 0,811 | 0,811 | 0,880 | 0,811 | 0,850 | 0,867 |
| Rank (1-10) | 0,801 | 0,811 | 0,522 | 0,808 | 0,877 | 0,892 | 0,811 | 0,811 | 0,880 | 0,811 | 0,846 | 0,874 |
| Rank (1-11) | 0,801 | 0,811 | 0,522 | 0,810 | 0,878 | 0,892 | 0,811 | 0,811 | 0,880 | 0,811 | 0,846 | 0,874 |
| Rank (1-12) | 0,801 | 0,811 | 0,493 | 0,846 | 0,889 | 0,918 | 0,799 | 0,811 | 0,886 | 0,811 | 0,875 | 0,888 |
| Rank (1-13) | 0,801 | 0,811 | 0,464 | 0,844 | 0,888 | 0,918 | 0,798 | 0,811 | 0,886 | 0,811 | 0,873 | 0,889 |
| Rank (1-14) | 0,801 | 0,811 | 0,559 | 0,853 | 0,897 | 0,919 | 0,798 | 0,811 | 0,887 | 0,811 | 0,882 | 0,897 |
| Rank (1-15) | 0,801 | 0,811 | 0,537 | 0,853 | 0,896 | 0,921 | 0,798 | 0,811 | 0,887 | 0,811 | 0,882 | 0,896 |
| Rank (1-16) | 0,801 | 0,811 | 0,598 | 0,852 | 0,896 | 0,921 | 0,800 | 0,811 | 0,888 | 0,811 | 0,879 | 0,887 |
| Rank (1-17) | 0,802 | 0,811 | 0,580 | 0,850 | 0,894 | 0,920 | 0,796 | 0,811 | 0,889 | 0,811 | 0,876 | 0,890 |
| Rank (1-18) | 0,802 | 0,811 | 0,580 | 0,847 | 0,894 | 0,920 | 0,796 | 0,811 | 0,889 | 0,811 | 0,876 | 0,890 |
| Rank (1-19) | 0,803 | 0,811 | 0,573 | 0,877 | 0,922 | 0,936 | 0,814 | 0,811 | 0,929 | 0,811 | 0,912 | 0,923 |
| Rank (1-20) | 0,803 | 0,811 | 0,556 | 0,884 | 0,930 | 0,941 | 0,813 | 0,811 | 0,932 | 0,811 | 0,913 | 0,928 |

**Analysis of Results**

Based on table 2, which shows the classification accuracy of different machine learning models using various feature subsets from a dental care provider classification study, here's a detailed analysis of the analysis of model performance across feature subsets:

- k-Nearest Neighbors (kNN): starts at an accuracy of 0,808 with the top feature and remains relatively stable, with a slight peak at 0,841 when using the top 3 features. However, it shows a decline when more features are added, suggesting that kNN might not benefit significantly from additional features beyond the top few.[60]

• Decision Tree (Tree): consistently maintains an accuracy of 0,811 across all feature subsets, indicating that the model's performance is not significantly affected by the number of features used. This suggests a simple decision boundary that doesn't gain complexity with more features.[61]

• Support Vector Machine (SVM): starts at 0,627 but shows a dramatic drop to 0,275 with the top 3 features, indicating poor performance with fewer features. Performance improves slightly as more features are included, peaking at 0,598 with 16 features, but overall, it struggles, suggesting difficulty in finding an effective hyperplane in this dataset.[62]

• Stochastic Gradient Descent (SGD): it begins at 0,811 and shows a slight decrease with fewer features but then increases to 0,884 with all 20 features. This indicates that SGD benefits from more features, likely due to its regularization capabilities which help in handling the dataset's complexity.[63]

• Random Forest: starts at 0,804, with a steady increase in accuracy as more features are added, reaching 0,93 with all features. This suggests that Random Forest leverages the ensemble approach effectively, reducing overfitting and capturing more complex patterns with additional features.[64]

• Neural Network: shows a consistent upward trend from 0,811 to 0,941 with all 20 features. This indicates that Neural Networks are particularly effective at capturing intricate feature interactions, with performance improving with more data complexity.[65]

• Naive Bayes: maintains a stable accuracy of 0,811 initially but shows a slight dip to 0,796 with 16-18 features, then recovers. This model might not handle feature interactions well but performs decently on this dataset.[66]

• Logistic Regression: consistently at 0,811 across all subsets, suggesting it finds a linear decision boundary that isn't improved by additional features, indicating the data might not be linearly separable or the model is too simple for the task.[67]

• Gradient Boosting: starts at 0,811 and improves significantly to 0,932 with all features, showing a similar trend to Random Forest but with slightly better performance, highlighting the strength of boosting algorithms in this context.[68]

• Constant Classifier: always at 0,811, which reflects the accuracy of predicting the majority class, providing a baseline for comparison.[69]

• CN2 Rule Inducer: starts at 0,811, with performance increasing to 0,913 with all features, indicating that rule-based methods can also capture complex patterns effectively when provided with more features.[70]

• AdaBoost: begins at 0,805 and increases to 0,928 with all features, showing a steady improvement with more features, which is typical for boosting methods as they focus on correcting errors from previous models.[71]

Key Insights:

• Feature Subset Performance: most models show improvement or stability as more features are included, with exceptions like SVM and Naive Bayes showing some decline or less improvement, which might suggest overfitting or issues with the feature relevance for these models.[72,73,74]

• Model Robustness: ensemble methods (Random Forest, Gradient Boosting, AdaBoost) and Neural Networks demonstrate robustness and improvement with more features, suggesting they handle the complexity of the data well.[75,76,77]

• Model Selection: for this dataset, Neural Networks, Gradient Boosting, and Random Forest are the top performers, achieving accuracies over 93 % when using all features, indicating they are well-suited for this classification task.[78,79,80]

• Implication for Feature Selection: the consistent improvement or stability with more features suggests that feature selection, while important, should be balanced with the capacity of certain models to handle high-dimensional data without overfitting.[81,82]

This analysis helps in understanding which models might be more effective for dental care provider classification and how feature selection impacts model performance.[83,84,85]

## DISCUSSION

The application of machine learning (ML) in the classification of dental care providers into standard rendering providers and safety net clinic (SNC) providers presents a nuanced landscape of challenges and opportunities, as evidenced by our study. The dataset, comprising 24 300 instances with 20 features from 2018, was notably challenging due to its high percentage of missing values (38,1 %) and significant class imbalance, which are typical of real-world healthcare datasets. These characteristics necessitate robust preprocessing techniques like median and mode imputation for handling missing data and SMOTE for addressing class imbalance, which were effectively employed in this study.

Our feature ranking analysis revealed that treatment-related metrics, such as TXMT_USER_CNT and TXMT_SVC_CNT, were the most predictive, indicating that the volume and type of treatments provided are critical in distinguishing between provider types. This finding is particularly insightful because it underscores the operational aspect of provider classification over demographic variables, which showed minimal impact. This suggests that the nature of service delivery, rather than who receives the service, is key in identifying SNC providers, who are crucial for serving underserved populations.

The performance of various ML models across different subsets of these ranked features provided several insights. The Neural Network model's superior performance (94,1 % accuracy with all features) highlights its capability to capture complex interactions within high-dimensional data, which is beneficial in healthcare contexts where interactions between variables can be non-linear and intricate. Similarly, ensemble methods like Random Forest and Gradient Boosting also performed exceptionally well, suggesting that combining multiple models can enhance decision-making robustness, especially in datasets with noise and missing data.

The stability observed in models like Decision Trees and Logistic Regression across feature subsets indicates that while these models might not capture complexity as well as neural networks, they provide a reliable baseline, which is valuable for initial assessments or when computational resources are limited. However, the poor performance of SVM, particularly with fewer features, suggests that linear separation might not be suitable for this dataset, emphasizing the need for models capable of handling non-linearity.

A notable trend was the aggravating performance of the model with more features, and especially for SGD, Random Forest, Neural Networks and Gradient Boosting. This trend conforms to the notion that, in a difficult classification task such as ours, having more features can offer more context that assists in discriminating against fine differences between classes. But this also brings doubts on model complexity and overfitting, especially when simpler models like kNN also showed a decline as they add more features.

These findings have implications beyond the realm of academic interest to making of policy and resource allocation in practice. Health policymakers who can make healthcare disparities worse, can do so by assisting and funding SNC providers – which are critical in reducing healthcare disparities – to improve accuracy in classifying these types of providers. It also suggests feature ranking in the model development for it to guide in focusing on variables that are the most informative, making this classification more efficient and effective.

## CONCLUSIONS

In conclusion, this study has demonstrated the effectiveness of machine learning in classifying dental care providers, with particular success in identifying safety net clinic providers crucial for underserved populations. The high accuracy achieved by Neural Networks, Gradient Boosting, and Random Forest models, especially when utilizing all available features, underscores the potential of advanced ML techniques in handling the complexities of healthcare data, including high dimensionality, missing values, and class imbalance.

The dominance of treatment-related metrics over demographic information in our feature ranking analysis provides a new perspective on what drives the classification of provider types. This suggests that operational metrics are more indicative of a provider's role in the healthcare ecosystem than demographic variables, which have traditionally been focal points in such classifications.

The study's findings have significant implications for healthcare equity. By accurately identifying SNC providers, stakeholders can better allocate resources, tailor policies, and design interventions to enhance service delivery to vulnerable populations. This could lead to a more equitable distribution of dental care services, potentially reducing gaps in access and improving health outcomes for those most in need.

Moreover, the methodological advancements in combining feature ranking with ML models offer a blueprint for similar studies in other healthcare domains where data quality issues are prevalent. This approach not only improves model performance but also ensures that the models are built on a foundation of the most relevant data, optimizing both computational resources and predictive power.

Future research could explore the integration of more sophisticated data imputation techniques or delve into the dynamics of feature interactions over time, considering that healthcare datasets often evolve. Additionally, extending this methodology to other types of healthcare providers or different healthcare systems could further validate and refine our findings, contributing to a broader understanding of provider classification in healthcare analytics. This study, therefore, not only enhances our immediate understanding but also sets a precedent for future applications of ML in improving healthcare delivery and policymaking.

## BIBLIOGRAPHIC REFERENCES

1. Wahed MA, Alqaraleh M, Alzboon MS, Al-Batah MS. Application of Artificial Intelligence for Diagnosing Tumors in the Female Reproductive System: A Systematic Review. Multidiscip. 2025;3:54.

2.  Alqaraleh M, Al-Batah M, Salem Alzboon M, Alzaghoul E. Automated quantification of vesicoureteral reflux using machine learning with advancing diagnostic precision. Data Metadata. 2025;4:460.

3.  Salem Alzboon M, Subhi Al-Batah M, Alqaraleh M, Alzboon F, Alzboon L. Guardians of the Web: Harnessing Machine Learning to Combat Phishing Attacks. Gamification Augment Real [Internet]. 2025 Jan;3:91. Available from: http://dx.doi.org/10.56294/gr202591

4.  Alqaraleh M, Salem Alzboon M, Subhi Al-Batah M, Solayman Migdadi H. From Complexity to Clarity: Improving Microarray Classification with Correlation-Based Feature Selection. LatIA [Internet]. 2025 Jan 1;3:84. Available from: https://latia.ageditor.uy/index.php/latia/article/view/84

5.  Alzboon MS, Subhi Al-Batah M, Alqaraleh M, Alzboon F, Alzboon L. Phishing Website Detection Using Machine Learning. Gamification Augment Real [Internet]. 2025 Jan 16;3:81. Available from: http://dx.doi.org/10.56294/gr202581

6.  Alqaraleh M, Salem Alzboon M, Mohammad SA-B. Optimizing Resource Discovery in Grid Computing: A Hierarchical and Weighted Approach with Behavioral Modeling. LatIA [Internet]. 2025 Jan 1;3:97. Available from: https://latia.ageditor.uy/index.php/latia/article/view/97

7.  Alqaraleh M, Salem Alzboon M, Subhi Al-Batah M. Real-Time UAV Recognition Through Advanced Machine Learning for Enhanced Military Surveillance. Gamification Augment Real [Internet]. 2025 Jan 1;3:63. Available from: https://gr.ageditor.ar/index.php/gr/article/view/63

8.  Wahed MA, Alqaraleh M, Alzboon MS, Subhi Al-Batah M. AI Rx: Revolutionizing Healthcare Through Intelligence, Innovation, and Ethics. Semin Med Writ Educ [Internet]. 2025 Jan 1;4:35. Available from: https://mw.ageditor.ar/index.php/mw/article/view/35

9.  Abdel Wahed M, Alqaraleh M, Salem Alzboon M, Subhi Al-Batah M. Application of Artificial Intelligence for Diagnosing Tumors in the Female Reproductive System: A Systematic Review. Multidiscip [Internet]. 2025 Jan 1;3:54. Available from: https://multidisciplinar.ageditor.uy/index.php/multidisciplinar/article/view/54

10.  Wahed MA, Alqaraleh M, Salem Alzboon M, Subhi Al-Batah M. Evaluating AI and Machine Learning Models in Breast Cancer Detection: A Review of Convolutional Neural Networks (CNN) and Global Research Trends. LatIA [Internet]. 2025 Jan 1;3:117. Available from: https://latia.ageditor.uy/index.php/latia/article/view/117

11.  Alzboon MS, Alqaraleh M, Al-Batah MS. Diabetes Prediction and Management Using Machine Learning Approaches. Data Metadata [Internet]. 2025; Available from: https://doi.org/10.56294/dm2025545

12.  Alqaraleh M, Al-Batah MS, Alzboon MS, Alzboon F, Alzboon L, Alamoush MN. Echoes in the Genome: Smoking's Epigenetic Fingerprints and Bidirectional Neurobiological Pathways in Addiction and Disease. Semin Med Writ Educ [Internet]. 2025; Available from: https://doi.org/10.56294/mw2024.585

13.  Alqaraleh M, Al-Batah MS, Alzboon MS, Alzboon F, Alzboon L, Alamoush MN. From Puffs to Predictions: Leveraging AI, Wearables, and Biomolecular Signatures to Decode Smoking's Multidimensional Impact on Cardiovascular Health. Semin Med Writ Educ [Internet]. 2025; Available from: https://doi.org/10.56294/mw2024.670

14.  Abuashour A, Salem Alzboon M, Kamel Alqaraleh M, Abuashour A. Comparative Study of Classification Mechanisms of Machine Learning on Multiple Data Mining Tool Kits. Am J Biomed Sci Res 2024 [Internet]. 2024;22(1):1. Available from: www.biomedgrid.com

15.  Mowafaq SA, Alqaraleh M, Al-Batah MS. AI in the Sky: Developing Real-Time UAV Recognition Systems to Enhance Military Security. Data Metadata. 2024;3:417.

16.  Alqaraleh M, Alzboon MS, Al-Batah MS. Skywatch: Advanced Machine Learning Techniques for Distinguishing UAVs from Birds in Airspace Security. Int J Adv Comput Sci Appl [Internet]. 2024;15(11). Available from: http://dx.doi.org/10.14569/IJACSA.2024.01511104

17.  Wahed MA, Alzboon MS, Alqaraleh M, Al-Batah M, Bader AF, Wahed SA. Enhancing Diagnostic Precision

in Pediatric Urology: Machine Learning Models for Automated Grading of Vesicoureteral Reflux. In: 2024 7th International Conference on Internet Applications, Protocols, and Services (NETAPPS) [Internet]. IEEE; 2024. p. 1–7. Available from: http://dx.doi.org/10.1109/netapps63333.2024.10823509

18.   Abdel Wahed M, Al-Batah M, Salem Alzboon M, Fuad Bader A, Alqaraleh M. Technological Innovations in Autonomous Vehicles: A Focus on Sensor Fusion and Environmental Perception [Internet]. 2024 7th International Conference on Internet Applications, Protocols, and Services (NETAPPS). IEEE; 2024 Nov. Available from: http://dx.doi.org/10.1109/netapps63333.2024.10823624

19.   Alzboon MS, Alqaraleh M, Wahed MA, Alourani A, Bader AF, Al-Batah M. AI-Driven UAV Distinction: Leveraging Advanced Machine Learning. In: 2024 7th International Conference on Internet Applications, Protocols, and Services (NETAPPS) [Internet]. IEEE; 2024. p. 1–7. Available from: http://dx.doi.org/10.1109/netapps63333.2024.10823488

20.   Wahed MA, Alzboon MS, Alqaraleh M, Halasa A, Al-Batah M, Bader AF. Comprehensive Assessment of Cybersecurity Measures: Evaluating Incident Response, AI Integration, and Emerging Threats. In: 2024 7th International Conference on Internet Applications, Protocols, and Services (NETAPPS) [Internet]. IEEE; 2024. p. 1–8. Available from: http://dx.doi.org/10.1109/netapps63333.2024.10823603

21.   Alzboon MS, Al-Shorman HM, Alka'awneh SMN, Saatchi SG, Alqaraleh MKS, Samara EIM, et al. The Role of Perceived Trust in Embracing Artificial Intelligence Technologies: Insights from Jordan's SME Sector. In: Studies in Computational Intelligence [Internet]. Springer Nature Switzerland; 2024. p. 1-15. Available from: http://dx.doi.org/10.1007/978-3-031-74220-0_1

22.   Wahed MA, Alzboon MS, Alqaraleh M, Ayman J, Al-Batah M, Bader AF. Automating Web Data Collection: Challenges, Solutions, and Python-Based Strategies for Effective Web Scraping. In: 2024 7th International Conference on Internet Applications, Protocols, and Services, NETAPPS 2024 [Internet]. IEEE; 2024. p. 1–6. Available from: http://dx.doi.org/10.1109/netapps63333.2024.10823528

23.   Al-Batah M, Salem Alzboon M, Alqaraleh M, Ahmad Alzaghoul F. Comparative Analysis of Advanced Data Mining Methods for Enhancing Medical Diagnosis and Prognosis. Data Metadata. 2024;3:465.

24.   Alqaraleh M. Enhancing Internet-based Resource Discovery: The Efficacy of Distributed Quadtree Overlay. In: Proceedings of the 3rd International Conference on Applied Artificial Intelligence and Computing, ICAAIC 2024. 2024. p. 1619–28.

25.   Suryawanshi A, Behera N. Prediction of wear of dental composite materials using machine learning algorithms. Comput Methods Biomech Biomed Engin. 2024;27(3):400–10.

26.   Suryawanshi A, Behera N. Prediction of mechanical properties of dental composite materials using machine learning algorithms. Materwiss Werksttech. 2023;54(11):1350–61.

27.   Song IS, Choi ES, Kim ES, Hwang Y, Lee KS, Ahn KH. Associations of Preterm Birth with Dental and Gastrointestinal Diseases: Machine Learning Analysis Using National Health Insurance Data. Int J Environ Res Public Health. 2023;20(3).

28.   Oguzhan A, Peskersoy C, Devrimci EE, Kemaloglu H, Onder TK. Implementation of machine learning models as a quantitative evaluation tool for preclinical studies in dental education. J Dent Educ. 2024;

29.   Sadegh-Zadeh SA, Bagheri M, Saadat M. Decoding children dental health risks: a machine learning approach to identifying key influencing factors. Front Artif Intell. 2024;7.

30.   Zanini LG, Rubira-Bullen IR, Nunes F de L. Segmentation and Classification of Dental Caries in Cone Beam Tomography Images Using Machine Learning and Image Processing. In: International Joint Conference on Biomedical Engineering Systems and Technologies. 2024. p. 428–35.

31.   Leth Rasmussen E, Have Musaeus M, Dahl MR, Løvschall H, Musaeus P. Enhancing dental and medical students' self-regulated learning through multiple choice questions: An evaluative study using machine learning. Tidsskr Læring og Medier. 2024;17(29).

32.  More PP. Automated Dental Cavity Detection Using Machine Learning. Int J Res Appl Sci Eng Technol. 2024;12(1):849-53.

33.  Farook TH, Ahmed S, Giri J, Rashid F, Hughes T, Dudley J. Influence of Intraoral Scanners, Operators, and Data Processing on Dimensional Accuracy of Dental Casts for Unsupervised Clinical Machine Learning: An in Vitro Comparative Study. Int J Dent. 2023;2023.

34.  Kang IA, Njimbouom SN, Kim JD. Optimal Feature Selection-Based Dental Caries Prediction Model Using Machine Learning for Decision Support System. Bioengineering. 2023;10(2).

35.  Toledo Reyes L, Knorst JK, Ortiz FR, Brondani B, Emmanuelli B, Saraiva Guedes R, et al. Early Childhood Predictors for Dental Caries: A Machine Learning Approach. J Dent Res. 2023;102(9):999-1006.

36.  Chen CC, Mondal K, Vervliet P, Covaci A, O'Brien EP, Rockne KJ, et al. Logistic Regression Analysis of LC-MS/MS Data of Monomers Eluted from Aged Dental Composites: A Supervised Machine-Learning Approach. Anal Chem. 2023;95(12):5205-13.

37.  Kwack DW, Park SM. Prediction of medication-related osteonecrosis of the jaw (MRONJ) using automated machine learning in patients with osteoporosis associated with dental extraction and implantation: a retrospective study. J Korean Assoc Oral Maxillofac Surg. 2023;49(3):135-41.

38.  Ytzhaik N, Zur D, Goldstein C, Almoznino G. Obstructive Sleep Apnea, Metabolic Dysfunction, and Periodontitis—Machine Learning and Statistical Analyses of the Dental, Oral, Medical Epidemiological (DOME) Big Data Study. Metabolites. 2023;13(5).

39.  Schuch HS, Furtado M, Silva GFDS, Kawachi I, Chiavegatto Filho ADP, Elani HW. Fairness of Machine Learning Algorithms for Predicting Foregone Preventive Dental Care for Adults. JAMA Netw Open. 2023;6(11):E2341625.

40.  Wang R, Hass V, Wang Y. Machine Learning Analysis of Microtensile Bond Strength of Dental Adhesives. J Dent Res. 2023;102(9):1022-30.

41.  Erkartal HŞ, Tatlı M, Secgin Y, Toy S, Duman BS. Gender Estimation with Parameters Obtained From the Upper Dental Arcade by Using Machine Learning Algorithms and Artificial Neural Networks. Eur J Ther. 2023;29(3):352-8.

42.  Dogan OB, Boyacioglu H, Goksuluk D. Machine learning assessment of dental age classification based on cone-beam CT images: a different approach. Dentomaxillofac Radiol. 2024;53(1):67-73.

43.  Alqaraleh M. Enhanced Resource Discovery Algorithm for Efficient Grid Computing. In: Proceedings of the 3rd International Conference on Applied Artificial Intelligence and Computing, ICAAIC 2024. 2024. p. 925-31.

44.  Al-Batah MS, Salem Alzboon M, Solayman Migdadi H, Alkhasawneh M, Alqaraleh M. Advanced Landslide Detection Using Machine Learning and Remote Sensing Data. Data Metadata [Internet]. 2024 Oct 7;1. Available from: https://dm.ageditor.ar/index.php/dm/article/view/419/782

45.  Al-Shanableh N, Alzyoud M, Al-Husban RY, Alshanableh NM, Al-Oun A, Al-Batah MS, et al. Advanced ensemble machine learning techniques for optimizing diabetes mellitus prognostication: A detailed examination of hospital data. Data Metadata. 2024;3:363.

46.  Al-Batah MS, Alzboon MS, Alzyoud M, Al-Shanableh N. Enhancing Image Cryptography Performance with Block Left Rotation Operations. Appl Comput Intell Soft Comput. 2024;2024(1):3641927.

47.  Alqaraleh M, Alzboon MS, Al-Batah MS, Wahed MA, Abuashour A, Alsmadi FH. Harnessing Machine Learning for Quantifying Vesicoureteral Reflux: A Promising Approach for Objective Assessment. Int J online Biomed Eng. 2024;20(11):123-45.

48.  Muhyeeddin A, Mowafaq SA, Al-Batah MS, Mutaz AW. Advancing Medical Image Analysis: The Role of Adaptive Optimization Techniques in Enhancing COVID-19 Detection, Lung Infection, and Tumor Segmentation.

LatIA [Internet]. 2024 Sep 29;2(74):74. Available from: https://latia.ageditor.uy/index.php/latia/article/view/74

49.    Putri AK, Alzboon MS. Doctor Adam Talib's Public Relations Strategy in Improving the Quality of Patient Service. Sinergi Int J Commun Sci. 2023;1(1):42–54.

50.    Al-Batah MS, Alzboon MS, Alazaidah R. Intelligent Heart Disease Prediction System with Applications in Jordanian Hospitals. Int J Adv Comput Sci Appl. 2023;14(9):508–17.

51.    Alzboon MS, Al-Batah MS. Prostate Cancer Detection and Analysis using Advanced Machine Learning. Int J Adv Comput Sci Appl. 2023;14(8):388–96.

52.    Alzboon MS, Bader AF, Abuashour A, Alqaraleh MK, Zaqaibeh B, Al-Batah M. The Two Sides of AI in Cybersecurity: Opportunities and Challenges. In: Proceedings of 2023 2nd International Conference on Intelligent Computing and Next Generation Networks, ICNGN 2023. 2023.

53.    Alzboon MS, Al-Batah M, Alqaraleh M, Abuashour A, Bader AF. A Comparative Study of Machine Learning Techniques for Early Prediction of Diabetes. In: 2023 IEEE 10th International Conference on Communications and Networking, ComNet 2023 - Proceedings. 2023. p. 1–12.

54.    Alzboon MS, Al-Batah M, Alqaraleh M, Abuashour A, Bader AF. A Comparative Study of Machine Learning Techniques for Early Prediction of Prostate Cancer. In: 2023 IEEE 10th International Conference on Communications and Networking, ComNet 2023 - Proceedings. 2023. p. 1–12.

55.    Alzboon MS, Qawasmeh S, Alqaraleh M, Abuashour A, Bader AF, Al-Batah M. Machine Learning Classification Algorithms for Accurate Breast Cancer Diagnosis. In: 2023 3rd International Conference on Emerging Smart Technologies and Applications, eSmarTA 2023. 2023.

56.    Alzboon MS, Al-Batah MS, Alqaraleh M, Abuashour A, Bader AFH. Early Diagnosis of Diabetes: A Comparison of Machine Learning Methods. Int J online Biomed Eng. 2023;19(15):144–65.

57.    Alzboon MS, Qawasmeh S, Alqaraleh M, Abuashour A, Bader AF, Al-Batah M. Pushing the Envelope: Investigating the Potential and Limitations of ChatGPT and Artificial Intelligence in Advancing Computer Science Research. In: 2023 3rd International Conference on Emerging Smart Technologies and Applications, eSmarTA 2023. 2023.

58.    Alzboon MS. Survey on Patient Health Monitoring System Based on Internet of Things. Inf Sci Lett. 2022;11(4):1183–90.

59.    Alzboon M. Semantic Text Analysis on Social Networks and Data Processing: Review and Future Directions. Inf Sci Lett. 2022;11(5):1371–84.

60.    Alzboon MS, Aljarrah E, Alqaraleh M, Alomari SA. Nodexl Tool for Social Network Analysis. Turkish J Comput Math Educ. 2021;12(14):202–16.

61.    Alomari SA, Salaimeh S Al, Jarrah E Al, Alzboon MS. Enhanced logistics information service systems performance: using theoretical model and cybernetics' principles. WSEAS Trans Bus Econ [Internet]. 2020 Apr;17:278–87. Available from: https://wseas.com/journals/bae/2020/a585107-896.pdf

62.    Alomari SA, Alzboon MS, Al-Batah MS, Zaqaibeh B. A novel adaptive schema to facilitates playback switching technique for video delivery in dense LTE cellular heterogeneous network environments. Int J Electr Comput Eng [Internet]. 2020 Oct;10(5):5347. Available from: http://ijece.iaescore.com/index.php/IJECE/article/view/16563

63.    Alomari SA, Alqaraleh M, Aljarrah E, Alzboon MS. Toward achieving self-resource discovery in distributed systems based on distributed quadtree. J Theor Appl Inf Technol. 2020;98(20):3088–99.

64.    Alzboon MS, Mahmuddin M, Arif S. Resource discovery mechanisms in shared computing infrastructure: A survey. In: Advances in Intelligent Systems and Computing. 2020. p. 545–56.

65.   Shawawreh S, Alomari SA, Alzboon MS, Al Salaimeh S. Evaluation of knowledge quality in the E -learning system. Int J Eng Res Technol. 2019;12(4):548-53.

66.   Alomari, Alzboon, Zaqaibeh, Al-Batah, Saleh Ali, Mowafaq Salem, Belal MS. An Effective Self-Adaptive Policy for Optimal Video Quality over Heterogeneous Mobile Devices and Network Discovery Services. Appl Math Inf Sci [Internet]. 2019 May;13(3):489-505. Available from: http://www.naturalspublishing.com/Article.asp?ArtcID=19739

67.   Banikhalaf M, Alomari SA, Alzboon MS. An advanced emergency warning message scheme based on vehicles speed and traffic densities. Int J Adv Comput Sci Appl. 2019;10(5):201-5.

68.   Alzboon MS. Internet of things between reality or a wishing - list : a survey. Int J Eng \& Technol. 2019;7(June):956-61.

69.   Al-Batah M, Zaqaibeh B, Alomari SA, Alzboon MS. Gene Microarray Cancer classification using correlation based feature selection algorithm and rules classifiers. Int J online Biomed Eng. 2019;15(8):62-73.

70.   Al Tal S, Al Salaimeh S, Ali Alomari S, Alqaraleh M. The modern hosting computing systems for small and medium businesses. Acad Entrep J. 2019;25(4):1-7.

71.   Alzboon MS, Alomari S, Al-Batah MS, Alomari SA, Banikhalaf M. The characteristics of the green internet of things and big data in building safer, smarter, and sustainable cities Vehicle Detection and Tracking for Aerial Surveillance Videos View project Evaluation of Knowledge Quality in the E-Learning System View pr [Internet]. Vol. 6, Article in International Journal of Engineering and Technology. 2017. p. 83-92. Available from: https://www.researchgate.net/publication/333808921

72.   Arif S, Alzboon MS, Mahmuddin M. Distributed quadtree overlay for resource discovery in shared computing infrastructure. Adv Sci Lett. 2017;23(6):5397-401.

73.   Mahmuddin M, Alzboon MS, Arif S. Dynamic network topology for resource discovery in shared computing infrastructure. Adv Sci Lett. 2017;23(6):5402-5.

74.   Mowafaq Salem Alzboon M. Mahmuddin ASCA. Challenges and Mitigation Techniques of Grid Resource Management System. In: National Workshop on FUTURE INTERNET RESEARCH (FIRES2016). 2016. p. 1-6.

75.   Al-Batah MS. Ranked features selection with MSBRG algorithm and rules classifiers for cervical cancer. Int J Online Biomed Eng. 2019;15(12):4.

76.   Al-Batah MS. Integrating the principal component analysis with partial decision tree in microarray gene data. IJCSNS Int J Comput Sci Netw Secur. 2019;19(3):24-29.

77.   Alzboon MS, Arif AS, Mahmuddin M. Towards self-resource discovery and selection models in grid computing. ARPN J Eng Appl Sci. 2016;11(10):6269-74.

78.   Al-Batah MS, Al-Eiadeh MR. An improved binary crow-JAYA optimisation system with various evolution operators, such as mutation for finding the max clique in the dense graph. Int J Comput Sci Math. 2024;19(4):327-38.

79.   Alzboon MS, Sintok UUM, Sintok UUM, Arif S. Towards Self-Organizing Infrastructure : A New Architecture for Autonomic Green Cloud Data Centers. ARPN J Eng Appl Sci. 2015;1-7.

80.   Kapoor S, Sharma A, Verma A, Dhull V, Goyal C. A Comparative Study on Deep Learning and Machine Learning Models for Human Action Recognition in Aerial Videos. Int Arab J Inf Technol [Internet]. 2023;20(4).

81.   Al-Batah MS. Testing the probability of heart disease using classification and regression tree model. Annu Res Rev Biol. 2014;4(11):1713-25.

82.   SalemAlzboon, Mowafaq and Arif, Suki and Mahmuddin, M and Dakkak O. Peer to Peer Resource Discovery Mechanisms in Grid Computing: A Critical Review. In: The 4th International Conference on Internet Applications, Protocols and Services (NETAPPS2015). 2015. p. 48-54.

83.   Al-Batah MS. Modified recursive least squares algorithm to train the hybrid multilayered perceptron (HMLP) network. Appl Soft Comput. 2010;10(1):236-44.

84.   Al-Batah MS, Al-Eiadeh MR. An improved discreet Jaya optimisation algorithm with mutation operator and opposition-based learning to solve the 0-1 knapsack problem. Int J Math Oper Res. 2023;26(2):143-69.

85.   Al-Oqily I, Alzboon M, Al-Shemery H, Alsarhan A. Towards autonomic overlay self-load balancing. In: 2013 10th International Multi-Conference on Systems, Signals and Devices, SSD 2013. Ieee; 2013. p. 1-6.

## FINANCING

This work is supported from Jadara University, Zarqa University, University of Petra, and Taibah University.

## CONFLICT OF INTEREST

The authors declare that the research was conducted without any commercial or financial relationships that could be construed as a potential conflict of interest.

## AUTHORSHIP CONTRIBUTION

*Conceptualization:* Muhyeeddin Alqaraleh, Mowafaq Salem Alzboon.
*Data curation:* Mowafaq Salem Alzboon, Mohammed Hasan Abu-Arqoub
*Formal analysis:* Mowafaq Salem Alzboon, Muhyeeddin Alqaraleh.
*Research:* Mohammad Subhi Al-Batah, Mowafaq Salem Alzboon.
*Methodology:* Rashiq Rafiq Marie, Mowafaq Salem Alzboon.
*Project management:* Mohammad Subhi Al-Batah, Mowafaq Salem Alzboon.
*Resources:* Mohammed Hasan Abu-Arqoub, Mohammad Subhi Al-Batah.
*Software:* Rashiq Rafiq Marie, Mohammad Subhi Al-Batah.
*Supervision:* Mohammad Subhi Al-Batah, Mohammed Hasan Abu-Arqoub.
*Validation:* Mowafaq Salem Alzboon, Muhyeeddin Alqaraleh.
*Display:* Mohammad Subhi Al-Batah, Mowafaq Salem Alzboon.
*Drafting - original draft:* Rashiq Rafiq Marie, Mohammed Hasan Abu-Arqoub.
*Writing:* Mohammad Subhi Al-Batah, Mohammed Hasan Abu-Arqoub, Muhyeeddin Alqaraleh.